\documentclass[letterpaper, 10 pt, conference]{ieeeconf}  

\IEEEoverridecommandlockouts                              
\usepackage[utf8]{inputenc}
\usepackage[T1]{fontenc}

\usepackage{todonotes}
\usepackage{graphicx} 
\usepackage{amsmath} 
\usepackage{amssymb}
\usepackage[font=footnotesize]{caption}
\usepackage{subcaption}
\usepackage{float}
\usepackage{placeins}
\usepackage{tabu}
\usepackage{hyperref}

\usepackage{graphics} 
\usepackage{amsmath} 
\usepackage{amssymb}  

\hypersetup{
    colorlinks=true,
    linkcolor=black,
    citecolor=black,
    urlcolor=black,
}

\author{Joshua Moser$^{1}$, Julia Hoffman$^{2}$, Robert Hildebrand$^{3}$, Erik Komendera$^{4}$%
\thanks{This material is based on work supported by NASA, Langley Research Center under the Research Cooperative Agreement No. NNL09AA00A awarded to the  National Institute of Aerospace. This material was also partially based on work sponsored by the Northrop Grumman Undergraduate Research Experience in Industrial \& Systems Engineering at Virginia Tech project.  Any opinions, findings, and conclusions or recommendations expressed in this material are those of the authors and do not necessarily reflect the views of the Northrop Grumman Corporation.}
 \thanks{$^1$J. Moser is a PhD student in Mechanical Engineering at the FASER Lab, Virginia Tech, 
 Blacksburg, VA {\tt\small joshnm7@vt.edu}}%
\thanks{$^2$J. Hoffman is with the Grado Department of Industrial and Systems Engineering, Virginia Tech, VA  24060, USA
        {\tt\small hjulia99@vt.edu}}
\thanks{$^3$R. Hildebrand is with Faculty of the Grado Department of Industrial and Systems Engineering, Virginia Tech, Blacksburg, VA  24060, USA
        {\tt\small rhil@vt.edu}}%
\thanks{$^4$E. Komendera is an Assistant Professor in Mechanical Engineering and Director of the FASER Lab, Virginia Tech, 
 Blacksburg, VA {\tt\small komendera@vt.edu}}%
}

\title{\LARGE \bf A Flexible Job Shop Scheduling Representation of the Autonomous In-Space Assembly Task Assignment Problem} 

\begin{document}

\maketitle

\begin{abstract}
As in-space exploration increases, autonomous systems will play a vital role in building the necessary facilities to support exploration. To this end, an autonomous system must be able to assign tasks in a scheme that efficiently completes all of the jobs in the desired project. This research proposes a flexible job shop problem (FJSP) representation to characterize an autonomous assembly project and then proposes both a mixed integer programming (MIP) solution formulation and a reinforcement learning (RL) solution formulation. The MIP formulation encodes all of the constraints and interjob dynamics a priori and was able to solve for the optimal solution to minimize the makespan. The RL formulation did not converge to an optimal solution but did successfully learn implicitly interjob dynamics through interaction with the reward function. Future work will include developing a solution formulation that utilizes the strengths of both proposed solution methods to handle scaling in size and complexity.
\end{abstract}

\section{Introduction} \label{sec:Introduction}
\begin{figure}
    \centering
    \includegraphics[width=\columnwidth]{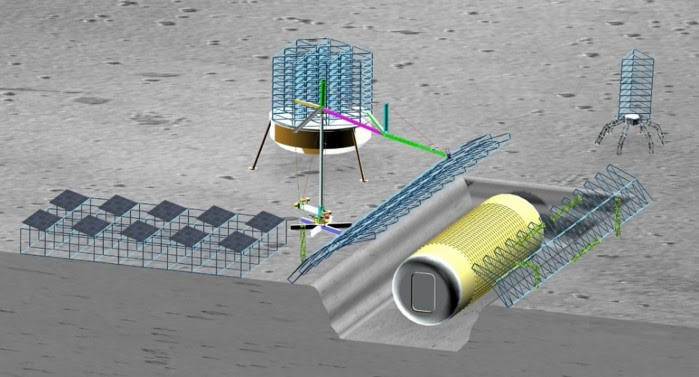}
    \caption{Concept of autonomous robot assembly for in-space facilities \cite{noauthor_virginia_nodate}.}
    \label{fig:inSpaceBase}
\end{figure}

As autonomous robotic systems are deployed to support efforts in industry or exploration it is vital that these systems have the ability to autonomously respond to new environments and off-nominal conditions. As in-space exploration continues to increase, with humanity returning to the moon and journeying to Mars according to the current directive for NASA, the required infrastructures will grow in scale and complexity. Constructing and maintaining these infrastructures with an astronaut workforce raises circular prerequisite errors given these infrastructures will be required to support the presence of astronauts. A solution to this integrates the use of autonomous robotic systems where these autonomous units collaboratively assemble and maintain infrastructures such as living quarters and power acquisition facilities similar to those in Fig. \ref{fig:inSpaceBase}, prior to the arrival of astronauts \cite{belvin_-space_2016}. Due to the cost of sending equipment into space and the time lost delivering it to a location such as Mars, it is not feasible to deliver pre-assembled facilities. This will lead to a variety of tasks requiring autonomous robotic attention. To ensure that a single robot's failure does not halt construction, there must be ability overlap across the different types of jobs that each robot is capable of completing. This overlap of ability between robotic units creates the possibility of many feasible assembly schemes. Solving for valid schemes that are also efficient will become more important as the number of assemblies increases. In addition to being solved before deployment, new schemes may need to be determined after work has already begun if an off-nominal occurrence causes the original scheme to be invalid.

This work seeks to provide a general problem formulation that describes the in-space assembly task assignment problem, facilitating the application of different solution methods that seek a valid and optimal assembly scheme. Additionally, two possible solution formulations were developed and evaluated. This general formulation takes the form of a flexible job shop scheduling problem (FJSP) which was then utilized in the investigation of two different solution methodologies, mixed integer programming (MIP) and reinforcement learning (RL). A MIP solution methodology was selected due to its extensive use in solving job shop scheduling problems (JSP) \cite{OZGUVEN2012846, ku_mixed_2016} and its ability to find provably optimal solutions for these problems given enough computation time. An RL approach was chosen due to its inherent ability to learn from elements not explicitly defined in the solution formulation. This ability allows for increasing interactions and prerequisites between jobs without additional complexity to the state space \cite{sutton_reinforcement_1998}. The rest of this paper is structured in the following way. Section \ref{sec:GeneralProblemFormulation} will discuss the general problem formulation as an FJSP. Section \ref{sec:ExerimentalScenario} will then define the realistic scenario used in the solution evaluations. Section \ref{sec:MIPFormulation} will discuss the MIP solution formulation to the FJSP followed by section \ref{sec:RLFormulation} which will discuss the RL solution formulation. Section \ref{sec:Simulation} will describe the simulation details for the MIP and RL formulations and section \ref{sec:Results} will discuss the results from both of these simulations. The final section, section \ref{sec:DiscussionConclusion}, will discuss the results and future research.  

\section{General Problem Formulation} \label{sec:GeneralProblemFormulation}
This paper proposes framing the autonomous assembly task assignment problem as a FJSP, an extension of the JSP, where each operation can be processed on any machine. In this formulation, a given job, $j \in J$, will have a set of operations, $O_{jp}$, where $p \in P$ is a processing plan defining the processing strategy. This processing strategy defines how the operations are processed by the machines (robotic units) represented as $m \in M$. Depending on the the type of operation, different types of robots will have different completion efficiencies. If $r \in R$ represents a type of robot working on a specific project (a set of jobs required to complete the desired facility or structure) and $q \in Q$ represents a type of operation in $J$ then the completion efficiency for each $(r, q)$ can be represented as an $|R|\times|Q|$ efficiency matrix, $\mathcal{E}$. In many projects, the work on a given job cannot begin until a different job in the project has already been completed. To efficiently represent this precedence constraint a directed acyclic graph (DAG), $G(V,\mathcal{A})$, is defined where the set of vertices, $V$, represent the jobs and the arcs, $\mathcal{A}$, represent the directed precedence paths between jobs. Additionally, $\mathcal{H}$ represents a set of these arcs that require machine-operation continuity.  In an assembly scenario, it is important to include information for the distance between different jobs in the task assignment process. These distances can be represented as the edges, $E$, of a completed graph, $G(V,E)$, where the vertices are the jobs in the project.  


\section{Experimental Scenario} \label{sec:ExerimentalScenario}
\begin{figure}[!b]
\hspace{-0.2cm}
  \begin{tabular}[b]{cc}
    \begin{tabular}[b]{c}
      \begin{subfigure}[b]{0.49\columnwidth}
        \includegraphics[width=1.1\columnwidth]{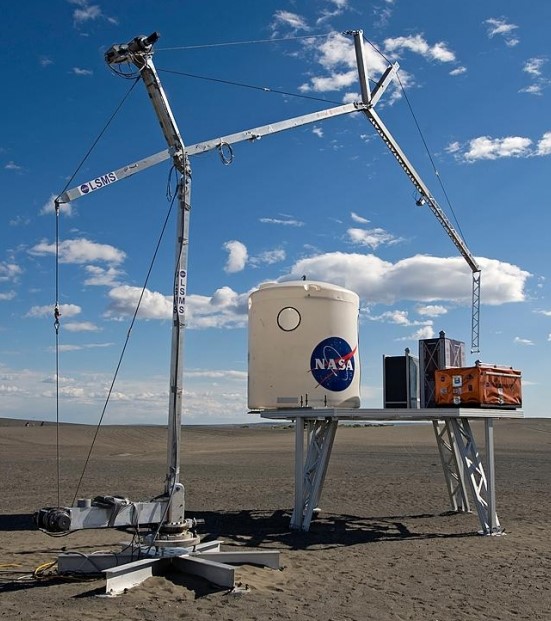}
        \caption{LSMS \cite{doggett_design_2008, dorsey_recent_2011}}
      \end{subfigure}\\
      \begin{subfigure}[b]{0.49\columnwidth}
        \includegraphics[width=1.1\textwidth]{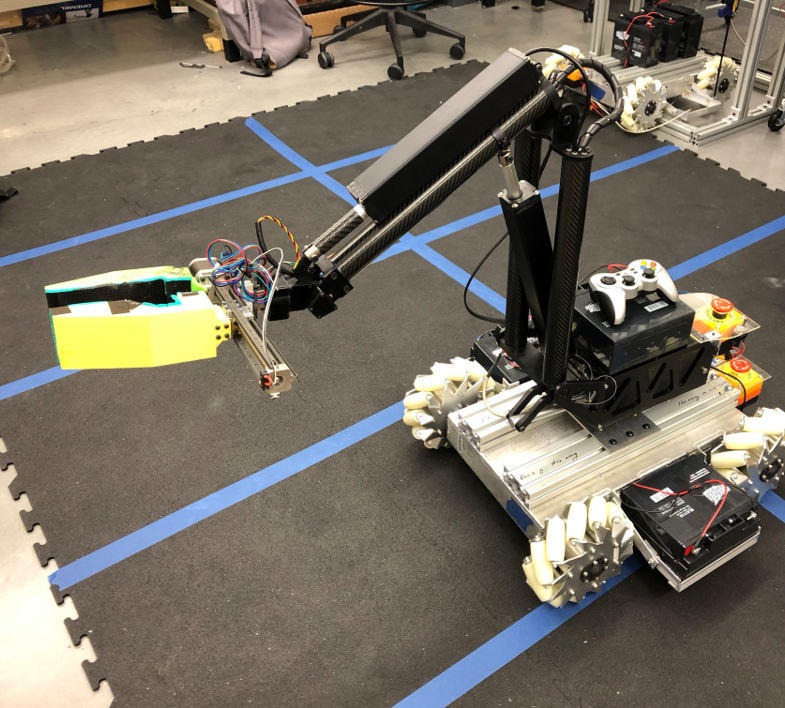}
        \caption{MARC \cite{noauthor_virginia_nodate}}
      \end{subfigure}
    \end{tabular}
    &
    \hspace{-0.5cm}
    \begin{subfigure}[b]{0.482\columnwidth}
    \centering
      \includegraphics[width=0.74\textwidth]{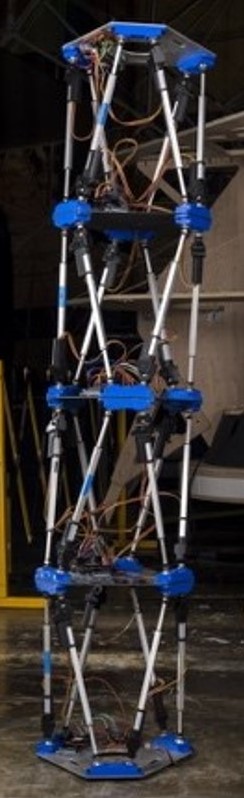}
      \caption{Assembler \cite{moser_reinforcement_2019}}
    \end{subfigure}
  \end{tabular}
  \caption{Robots considered for autonomous solar assembly problem and simulation. \label{fig:Robots}}
\end{figure}

To illustrate the implementation of this formulation and provide a realistic experimental example for simulation, consider a solar farm assembly scenario where robotic systems must autonomously assemble solar panels for an in-space application on the surface of the moon. For a mission such as this, it is beneficial to launch the solar panels as base components to maximize the volume usage in the delivery system. These components will then be deposited in a storage area on the edge of the project's workspace. As mentioned in section \ref{sec:Introduction}, robotic units with a range of abilities would be present to assemble these base components into solar panels. One such unit would be a large robot, capable of unloading the basic components from a lander or moving the components quickly across the workspace. NASA Langley Research Center (LaRC) has developed the Lightweight Surface Manipulation System (LSMS) \cite{doggett_design_2008, dorsey_recent_2011} to accomplished this type of task. This assembly scenario will also require robots that are mobile and capable of utilizing grippers to: move assembly components, operate tools used for affixing components, and collaborate in the carrying of larger components. This type of robot will likely take a form akin to the rover and arm planned in the Mars 2020 mission \cite{williford_chapter_2018} where a robotic arm is affixed to a rover chassis capable of traversing the terrain in the assembly workspace. The experimental robot fulfilling this role in the research presented here is the Mobile Assembly Robotic Collaborator (MARC) at the Field and Space Experimental Robotics (FASER) Laboratory \cite{noauthor_virginia_nodate}. In autonomous assembly, precise manipulation is required to align components before they are permanently affixed into place. To accomplish this type of task, LaRC has developed the Assembler robot, a serialized parallel robot consisting of modularizable Stewart platforms \cite{moser_reinforcement_2019}. Built units of these robots can be seen in Fig. \ref{fig:Robots}. The robots used in the following simulations are modeled after these robots. 

\begin{figure}[!t]
  \begin{tabular}[b]{cc}
    \begin{tabular}[b]{c}
      \begin{subfigure}[b]{0.45\columnwidth}
        \includegraphics[width=1.2\textwidth]{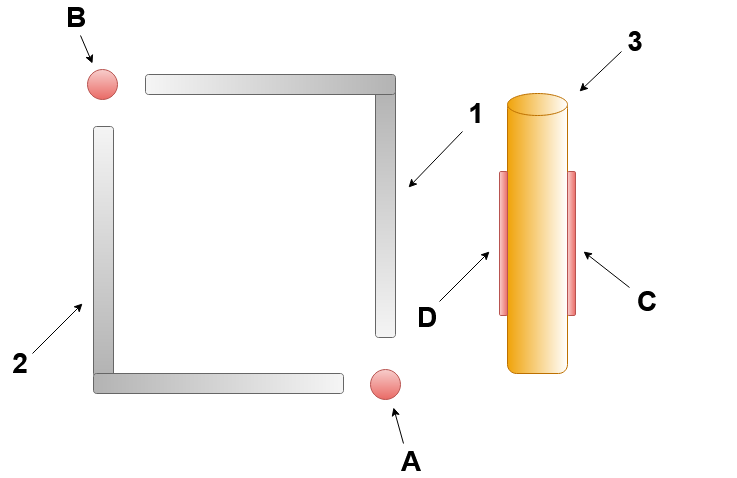}
        \caption{Solar array components \label{fig:SolarArrayComponents}}
      \end{subfigure}\\
      \begin{subfigure}[b]{0.45\columnwidth}
        \includegraphics[width=\textwidth]{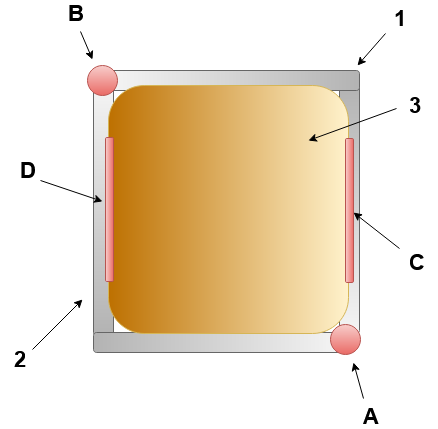}
        \caption{Solar array assembled \label{fig:SolarArrayAssembled}}
      \end{subfigure}
    \end{tabular}
    &
    \begin{subfigure}[b]{0.45\columnwidth}
    \centering
      \includegraphics[width=0.695\textwidth]{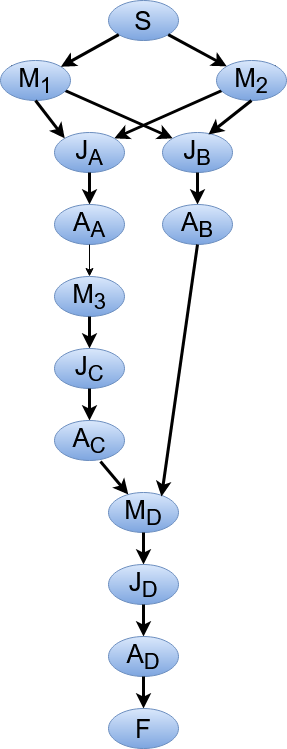}
      \caption{Solar array DAG \label{fig:SolarArrayDAG}}
    \end{subfigure}
  \end{tabular}
  \caption{Components diagram and DAG for solar array assembly project. \label{fig:SolarArrayJobs}}
\end{figure}

As a representation of the solar panel assembly, this work will model the solar panel as three components; two for the frame and one for the solar cell sheet, shown in Fig. \ref{fig:SolarArrayComponents} as $1, 2, \& \; 3$. These three components will be attached via the connection points $A, B, C, \& \; D$ shown in the same figure. 
To assemble this solar array, the components will need to be moved, jigged (aligned), and then affixed (welded or attached) at the connection points. These three types of jobs, represented by $M, J, \& \; A$ respectively, will have operations consisting of combinations of four operation types that constitutes the set $Q$. The operation types and the set of operations for each job are given in Tables \ref{tab:OpTypes} $\&$ \ref{tab:OpsInJobs} respectively. The completion efficiency for each robot type to each operation type is also given in Table \ref{tab:AbilityTable}. To successfully assemble the solar panel, both of the frame components must first be moved to the assembly location. Following this, each component of the frame must be aligned into position with respect to the adjoining piece (for this representation only one component needs to be aligned to a stationary piece for each frame piece). Once aligned, it must be welded into place before the alignment can be released. When the frame is ready to receive the solar cell sheet, it is moved into place. One end is affixed to the frame and then the cell is unrolled before it is affixed to the final side of the frame ending the assembly sequence. The DAG representing these precedence constraints for assembling a solar panel is given in Fig. \ref{fig:SolarArrayDAG} and the holding constraints are between the jigging jobs and their respective affixing jobs since the components can not be disturbed between these jobs. This representation of the solar farm assembly problem will be used in the following sections demonstrating the solution formulations to solve for a valid assembly scheme.
\begin{table}[!b]
    \centering
      \caption{Operation types and completion efficiencies. \label{tab:OpTypes}\label{tab:AbilityTable}}
      \resizebox{0.9\columnwidth}{!}{%
        %
        \begin{tabular}{|c|c|c|c|c|c|c|c|}  
        \hline
        q & \textbf{Description} & Assembler & MARC & LSMS\\
        \hline
        0 & Hold frame link & 1 & 5 & 10 \\
        1 & Hold solar cell sheet & 1 & 5 & 10 \\
        2 & Weld connection point & 5 & 1 & 10 \\
        3 & Locomote & 100 & 5 & 1 \\
        \hline
        \end{tabular}
        }
\end{table}
\begin{table}[]
    \centering
    \caption{Jobs, Process plans, and operation lists.
    In this example, each process job has exactly one process plan. For example, job $A_A$ has one process plan $O_{A_a 0}$ with operations three operations of types $0,0$, and $2$. \label{tab:OpsInJobs}}
    \resizebox{0.8\columnwidth}{!}{%
        \begin{tabular}{|c|c|l|}
            \hline
             Jobs $J$ & Process plans $P_j = \{O_{j0},\dots, O_{jn_{j}}\}$ \\
            \hline
            $M_1$ & $\{[3]\}$ \\
         $M_2$ & $\{[3]\}$ \\
             $J_A$ & $\{[0, 0]\}$ \\
             $J_B$ & $\{[0, 0]\}$ \\
             $A_A$ & $\{[0, 0, 2]\}$ \\
             $A_B$ & $\{[0, 0, 2]\}$ \\
             $M_3$ & $\{[3]\}$ \\
             $J_C$ & $\{[1]\}$\\
             $A_C$ & $\{[1, 2]\}$ \\
             $M_D$ & $\{[3]\}$ \\
             $J_D$ & $\{[0, 1]\}$ \\
             $A_D$ & $\{[1, 2]\}$ \\
            \hline
        \end{tabular}}
\end{table}

\section{Mixed Integer Programming Formulation}
\label{sec:MIPFormulation}

Mixed integer programming is a powerful tool for solving optimization problems and provides a way to compare solutions to calculated bounds on optimal solutions.  In particular, the backbone of commercial solvers use \textit{Branch \& Bound} to intelligently explore the space of feasible solutions while producing tighter bounds on how good the optimal solution could be.  As a result, a report on a measure of optimality gap (a ratio of how far the best found solution is from optimality) can be given. The formulation originally inspired by \cite{OZGUVEN2012846} is written to allow multiple process plans per job, allowing for different types of operations to be done to complete a job, and hence potential for faster job completion.   

The model is described in the following form. For shorthand, $\forall j, p, o, m$ means $\forall j \in J$, $p \in P_j$, $o \in O_{jp}$, $m \in M$, respectively.  Furthermore, a specific process plane is referred to as $jp$ and operations as $jpo$, to reference which job or plan they come from.  

\vspace{0.1cm}
\underline{Sets}\\
 \begin{tabular}{ll}
 	$\mathcal A \subseteq J \times J$ & arc set for precedence of jobs\\
 	$\mathcal H\subseteq J \times J$ & operation continuity set \end{tabular}
 	
\vspace{0.2cm}
\underline{Parameters}\\
\begin{tabular}{ll}
	$T_{jpom}$ &  Operation time of $m$ to do $jpo$\\
	$S_{j j' m}$ & Setup time of $m$ moving from $j$ to $j'$
\end{tabular}

\vspace{0.2cm}
\underline{Continuous Variables} (non-negative)\\
\begin{tabular}{ll}
	  $s_{j}$  &  the start time of job $j$\\
	  $c_{j}$  &   completion time of job $j$\\
	  $c_{max}$  &   upper bound of completion times\\
	  $c_{jpo}$  & completion time of operation $jpo$ \\
	  $c_{jm}$  & completion time of machine $m$ on job $j$
	 \end{tabular}
	 
\vspace{0.2cm}
\underline{Binary Variables}\\	 
	 \begin{tabular}{ll}
	  $x_{jp}$ &  plan $p$ is selected for job $j$\\
	  $x_{jm} $ & machine $m$ is selected for job $j$\\
	  $x_{jj'm} $ & machine $m$ moves from $j$ to $j'\neq j$\\
	  $x_{jpom}$ &  machine $m$ assigned to operation $jpo$\\
\end{tabular}

\vspace{0.2cm}
\underline{Mixed Integer Programming Model}
\begin{subequations}

\textit{Objective function:}
\begin{align}
   \min \ \ \  & c_{\max}
\end{align}
\textit{Completion and start time constraints:}
The total makespan $c_{\max}$ occurs when all jobs are finished, and jobs are finished when all operations for the job are finished.
Furthermore, from the DAG, jobs cannot start until preceding jobs are completed:  
\begin{align}
   c_{\max} &\geq c_{j} \geq c_{jpo} & \forall j,p,o\\
   s_j &\geq c_{j'} & \forall (j',j) \in \mathcal A
\end{align}
If $m$ moving from $j'$ to $j$, then $j$ must start after completion of $j'$ and travel time of $m$ to $j$: ($L$ is a large enough number)
\begin{align}
    s_{j} &\geq  c_{j'} + (S_{jj'm}+L)x_{jj'm} - L & \forall j \neq j', m
    \end{align}
    \begin{align}
    c_{j} &\geq s_{j} + T_{jpom} x_{jpom} & \forall j,p,o,m
    \end{align}
    
\textit{Assignment constraints:}
\noindent
Exactly 1 process plan chosen for job $j$:
\begin{align}
    \sum_{p \in P_j} x_{jp} &= 1&  \forall j
\end{align}
If m assigned to operation $jpo$, then process plan p must be chosen for job $j$:
\begin{align}
    x_{jpom} & \leq x_{jp} & \forall j,p,o,m
    \end{align}
    If $jp$ is chosen, then each  $o$ in $jp$ must be assigned exactly 1 machine:
       \begin{align}
    x_{jp} & = \sum_{m \in M} \sum_{o \in O_{jp}} x_{jpom} & \forall j,p
    \end{align}
    If $m$ assigned to $jpo$, then $p$ must be chosen for $j$:
    \begin{align}
    \sum_{p \in P_j} \sum_{o \in O_{jp}}
    x_{jpom} &\leq x_{jm} & \forall j,p,m
    \end{align}
    If machine $m$ chosen for job $j$, then it must have exactly one edge entering and leave that job on its path:
    \begin{align}
    x_{jm} & = x^{\text{end}}_{jm} + \sum_{j'\in J\setminus\{j\}} x_{jj'm} & \forall j,m \\
    x_{jm} &= x^{\text{start}}_{jm} + \sum_{j' \in J\setminus\{j\}} x_{j'jm} & \forall j,m
    \end{align}
    $m$ must have path from starting to ending node:
    \begin{align}
    \sum_{j \in J} x^{\text{start}}_{jm} =
    \sum_{j \in J} x^{\text{end}}_{jm} &= 1 & \forall m
\end{align}
\textit{Operation continuity:}
for specific arcs  $(j, j') \in \mathcal{H}$, if machine $m$ does operation $\bar o$ for $j$, then it must do operation $\bar o$ for $j'$ and it must transition to job $j'$ next
\begin{align}
    \sum_{p \in P_j} x_{jp\bar o m} &= \sum_{p' \in P_j} x_{j'p'\bar o m}  &\forall m, (j,j') \in \mathcal{H},\\
    x_{jj'm} &\geq \sum_{p \in P_j} x_{jp\bar o m}  & \forall m, (j,j') \in \mathcal{H}.
\end{align}
 

\end{subequations}
\vspace{0.5cm}
\section{Reinforcement Learning Formulation}\label{sec:RLFormulation}
As stated previously, a reinforcement learning method was chosen as a solution method in an attempt to implicitly learn some of the environment dynamics that were explicitly encoded in the mixed integer programming formulation. The chosen reinforcement learning algorithm for this evaluation was an off-policy temporal difference algorithm known as Q-learning. The Q-learning equation is defined by
\begin{multline} \label{eq:QLearning}
    \mathcal{Q}(s_t, a_t) \leftarrow \mathcal{Q}(s_t, a_t) +  \\ \alpha \left[r_{t+1} + \gamma \; \underset{a}{\max} \; \mathcal{Q}(s_{t+1}, a) - \mathcal{Q}(s_t, a_t) \right]. 
\end{multline}

This temporal difference method samples the environment, learning from experiences rather then a dynamics model, giving it similarities to a Monte Carlo Method while updating estimates based partially on previously learned estimates making it akin to a dynamics programming method \cite{sutton_reinforcement_1998}. This method updates $\mathcal{Q}$ (the action-value function) to approximate the optimal action-value function off-policy (i.e. independent of the current policy). The discount factor, $\gamma$, controls the weight of the future effect of the action chosen at time $t$. The learning rate, $\alpha$, controls the weight at which the present value of the action-value entry is updated from the temporal difference evaluation. To evaluate if the RL was able to learn the precedent and holding constraints the action space was the same size as the state space, allowing the policy to choose any state as the next state. To minimize the total makespan a small penalty was incurred at every time step and a large penalty was given if the policy violated the precedent or holding constraints or if it tried to complete a job that had already been completed. 

\section{Simulation}\label{sec:Simulation}
\begin{figure}[b]
    \centering
    \includegraphics[width=1\columnwidth]{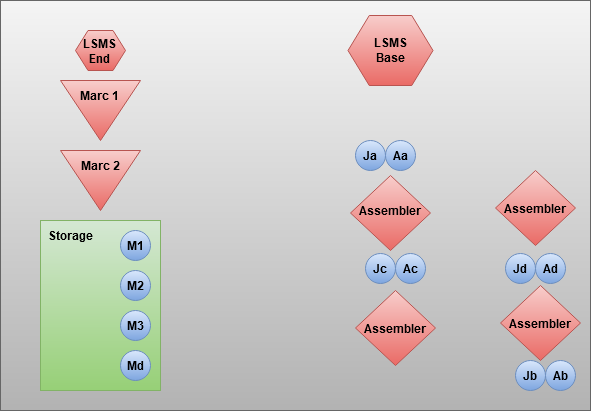}
    \caption{Project workspace used in the simulations.}
    \label{fig:workspace}
\end{figure}

To evaluate the solution methodologies with the scenario described above, the completion efficiency values shown in Table \ref{tab:AbilityTable} where chosen to reflect the general ability differences of the robots described above. The general spatial layout of this simulation environment is qualitatively shown in Fig. \ref{fig:workspace}. In this workspace, the Assembler robot is used in its disassociated state. That is, each Stewart Platform is spatially separated from the others while still being treated as a single robot unit. This is represented by the four red diamonds. Additionally, the LSMS base and its end-effector are denoted by the red hexagons while the MARC units are represented by the red triangles. The blue circles represent the qualitative spatial location of the jobs. In the simulations the spatial distances between jobs are shown in Table \ref{tab:JobDist}. The time it takes a robot to traverse this distance is the distance value multiplied by the specific robot's ability to locomote divided by a constant value ($30$ for all robots aside from MARC 2 which used a constant value of $25$).
\subsection{Mixed Integer Programming}
For the mixed integer programming solution simulation, the setup time was the amount of time it took a robot to traverse the distance between jobs. The current model formulation allows robots to begin the simulation at any location, which is easily adjusted by changing the fix starting points.  
For this simulation only one process plan was used, which prescribed one machine per task for a given job. The operation time was the completion efficiency of the robot assigned to work on it given in Table \ref{tab:AbilityTable}. Using these parameters and the formulation described in section \ref{sec:MIPFormulation} this simulation was evaluated using the Gurobi solver \cite{noauthor_gurobi_nodate}. The results from this simulation will be discussed in section \ref{sec:Results}.
\begin{table}
    \centering
    \caption{Spatial distances between jobs used in the simulations. \label{tab:JobDist}}
    \resizebox{0.9\columnwidth}{!}{%
    \begin{tabular}{|c|c|c|c|c|c|c|c|c|c|c|c|c|}
    \hline
     \textbf{Jobs} & M1 & M2 & Ja & Jb & Aa & Ab & M3 & Jc & Ac & Md & Jd & Ad \\
    \hline
     M1 & 0 &     0 &   43 &   50 &   38 &   55 &   0 &    36 &   35 &    0 &   56 &  55 \\
     M2 & 0 &     0 &   41 &   50 &   36 &   55 &    0 &   35 &   35 &    0 &   55 &  55 \\
    Ja & 43 &    41 &    0 &    0 &    0 &   0 &    40 &    0 &   0 &   40 &    0 &   0 \\
    Jb & 50 &    50 &    0 &    0 &    0 &   0 &    52 &    0 &    0 &   53 &    0 &   0 \\
    Aa & 38 &    36 &    0 &    0 &    0 &   0 &    35 &    0 &    0 &   35 &    0 &   0 \\
    Ab & 55 &    55 &    0 &    0 &    0 &   0 &    57 &    0 &    0 &   58 &    0 &   0 \\
     M3 & 0 &     0 &   40 &   52 &   35 &   57 &    0 &    35 &  35 &    0 &   55 &  55 \\
    Jc & 36 &    35 &    0 &    0 &    0 &   0 &    35 &     0 &   0 &   35 &    0 &   0 \\
    Ac & 35 &    35 &    0 &   0 &    0 &  0 &    35 &    0 &   0 &   35 &    0 &   0 \\
     Md & 0 &     0 &   40 &   53 &   35 &   58 &    0 &    35 &  35 &    0 &   55 &  55 \\
    Jd & 56 &    55 &    0 &   0 &    0 &   0 &    55 &    0 &    0 &   55 &    0 &   0 \\
   Ad &  55 &    55 &    0 &    0 &    0 &   0 &    55 &    0 &    0 &   55 &    0 &   0 \\
    \hline
    \end{tabular}
    }
\end{table}
\subsection{Reinforcement Learning}
The reinforcement learning simulation used the same setup and completion efficiency data as the MIP. However, due to the limitations stemming from the size of the state space, only a subset of the jobs, ${M_2, J_A, A_A, J_B, A_B}$, were used. These five jobs were chosen because they reflect the same constraints seen in the simulation for the MIP formulation. The state space for this Q-learning consisted of the five different jobs multiplied by the 24 possible different permutations based off the process plan yielding a state space of 120 entries, where each state is a job and process plan assignment combination. This leads to a state-action $\mathcal{Q}$-table of the size $120 \times 120$ since the action space consists of the next job and process plan to proceed to from a given state.  After a coarse grid search $\gamma = 0.4$ and $\alpha = 0.5$ were chosen for the learning parameters. The penalty for violating a constraint or trying to complete a job already completed was $-1E4$ whereas a the penalty for time unit spent was $-1$. This RL simulation was programmed using OpenAI's Gym framework \cite{noauthor_gym_nodate}.

\section{Results} \label{sec:Results}
\subsection{Mixed Integer Programming}
\label{sec:MIP-results}
\begin{figure}[]
    \centering
    \includegraphics[width = \columnwidth]{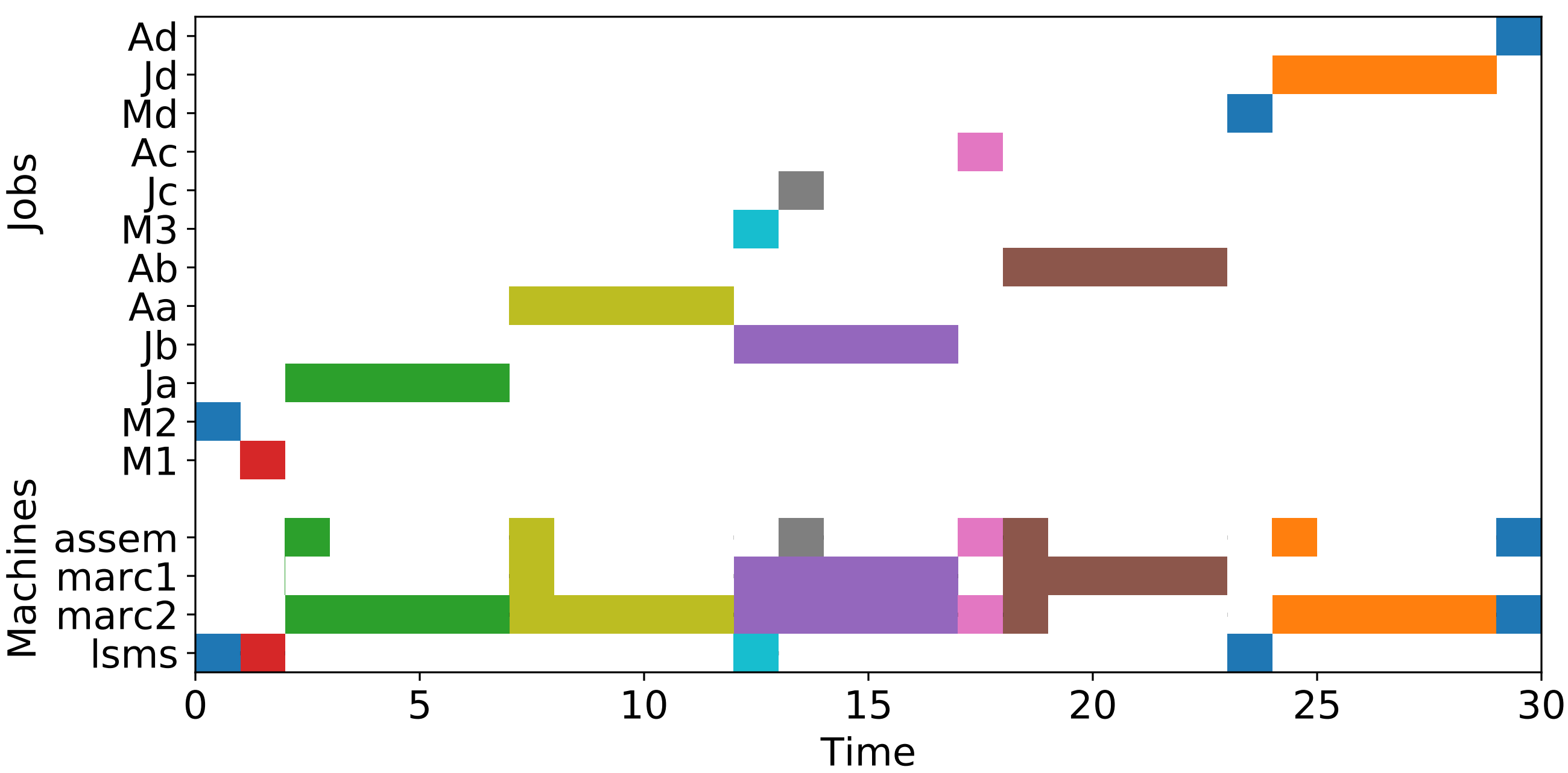}
    \caption{Schedule of machines and jobs over time for one solar panel from MIP. Note that the choice of starting position allows for a solution where machines do not need to move anywhere.}
    \label{fig:MIPBarGraph}
\end{figure}

\begin{figure}[]
    \centering
    \includegraphics[width = \columnwidth]{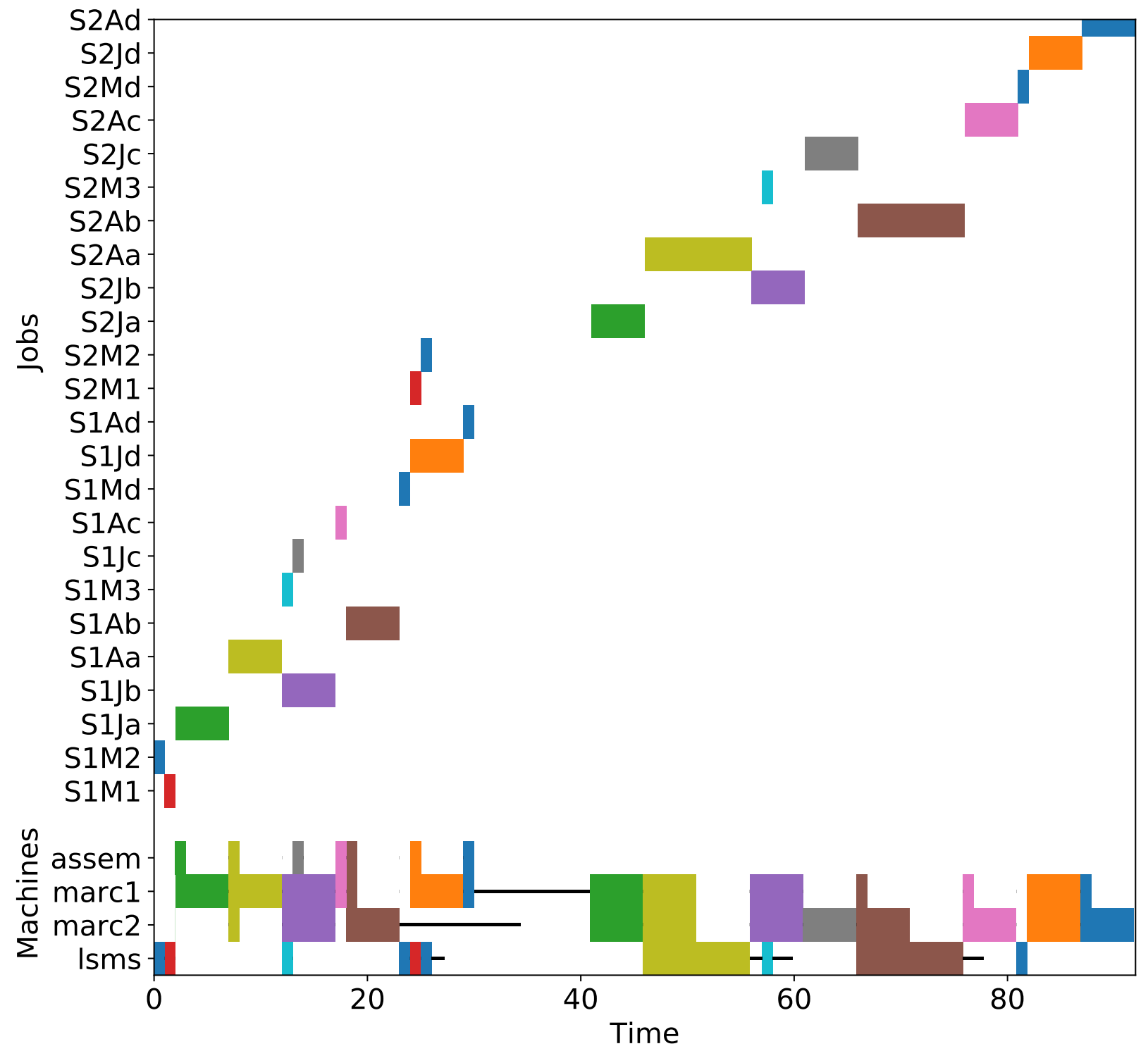}
    \caption{Schedule of machines and jobs over time with two solar panels from MIP. Distance between jobs on each solar panel was preserved, but the setup of the second solar panel is shifted a distance away from the first one.}
    \label{fig:MIPBarGraph2}
    \vspace{-0.3cm}
\end{figure}

The mixed integer program was solved using Gurobi version 9.0 and its Python API \cite{noauthor_gurobi_nodate}. Computations were done on a 2014 MacBook Air laptop running macOS Mojave with a 1.7 GHz Intel Core i7, and memory of 8 GB, 1600 MHz DDR3. For the single solar panel example, the MIP finds an optimal solution within 2 seconds and proves optimality in 5 seconds. A larger example with 2 solar panels was also tested where an optimal solution was found in 26 seconds, however, the solver had trouble proving optimality. The optimal schedule output from the MIP for both cases are shown in Figs.~\ref{fig:MIPBarGraph} and \ref{fig:MIPBarGraph2}. Colored rectangles indicate intervals of time that a job is being worked on, while black lines indicate a machine traveling to a different job.  Some jobs are located in the same place and do not require travel time between those jobs.

\subsection{Reinforcement Learning}
\begin{figure}[]
    \includegraphics[width = \columnwidth]{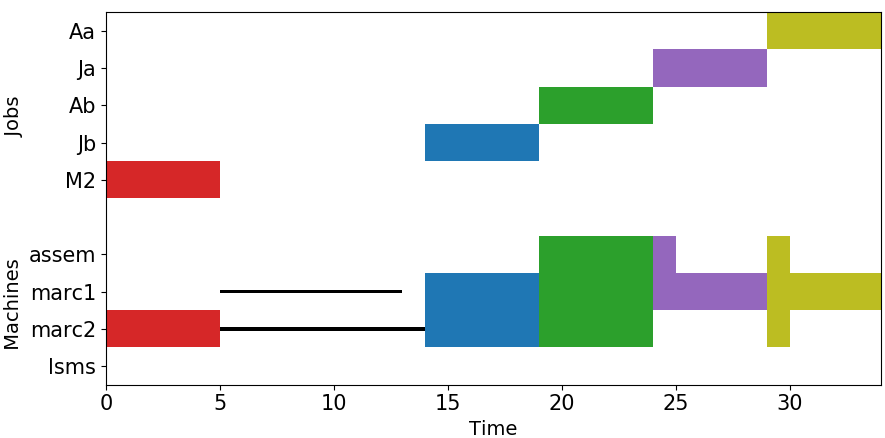}
    \caption{Schedule of machines and jobs over time for a reduced job set from RL. Note that the RL did not learn to use the LSMS to complete $M_2$ which would have been more efficient.}
    \label{fig:RLResults}
\end{figure}

\begin{figure}[]
\centering
  \begin{tabular}[b]{cc}
        \begin{tabular}[b]{c}
          \begin{subfigure}[b]{0.45\columnwidth}
            \includegraphics[width=0.8\columnwidth]{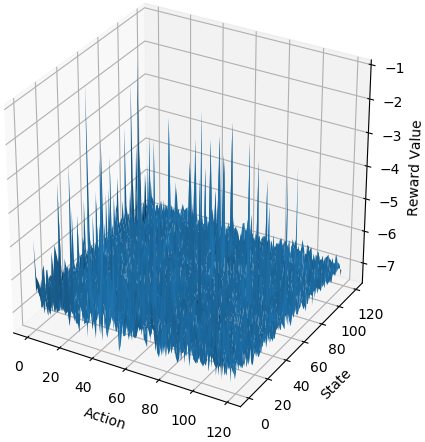}
            \caption{State-action Space for RL model. \label{fig:ActionStateSpaceRL}}
          \end{subfigure}\\
          \begin{subfigure}[b]{0.45\columnwidth}
            \includegraphics[width=0.8\columnwidth]{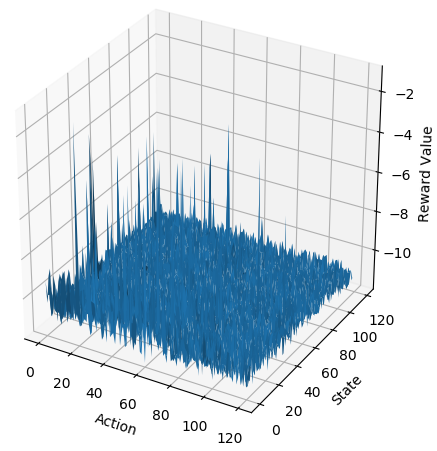}
            \caption{State-action space for alternative RL model. \label{fig:ActionStateAltModel}}
          \end{subfigure}\\
            \begin{subfigure}[b]{0.45\columnwidth}
            \includegraphics[width=0.8\columnwidth]{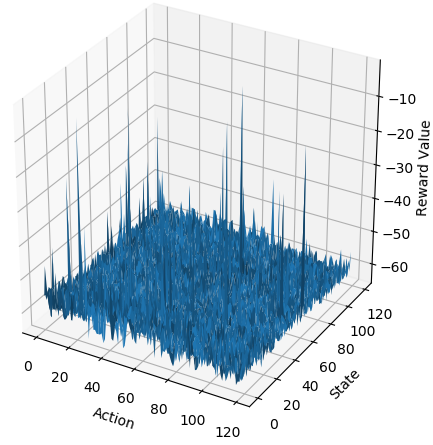}
            \caption{State-action space for second alternative RL model. \label{fig:ActionStateAltModel2}}
          \end{subfigure}
        \end{tabular}
    &
        \hspace{-0.4cm}
        \begin{tabular}[b]{c}
          \begin{subfigure}[b]{0.45\columnwidth}
            \hspace{-0.5cm}
            \includegraphics[width=1.1\columnwidth]{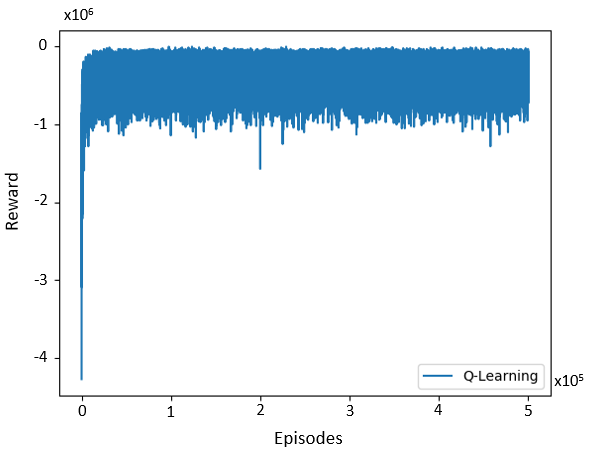}
            \caption{Learning curve for RL model. \label{fig:LearningCurveRL}}
          \end{subfigure}\\
          \begin{subfigure}[b]{0.45\columnwidth}
            \hspace{-0.5cm}
            \includegraphics[width=1.1\columnwidth]{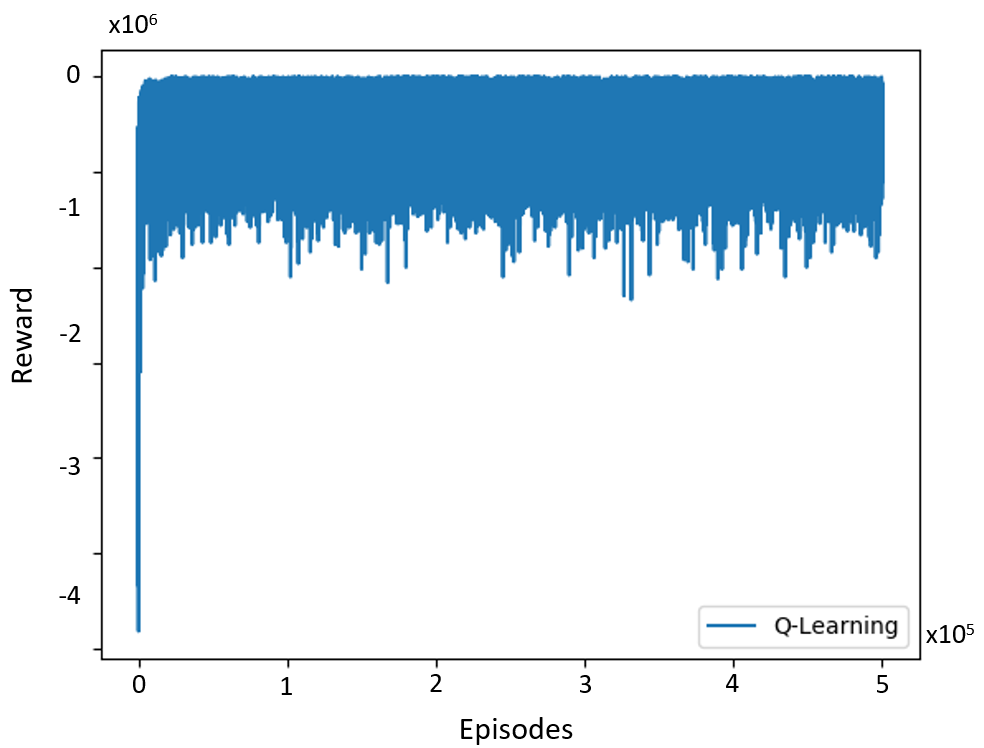}
            \caption{Learning curve for alternative RL model. \label{fig:LearningCurveAltModel}}
          \end{subfigure}\\
          \begin{subfigure}[b]{0.45\columnwidth}
            \hspace{-0.5cm}
            \includegraphics[width=1.1\columnwidth]{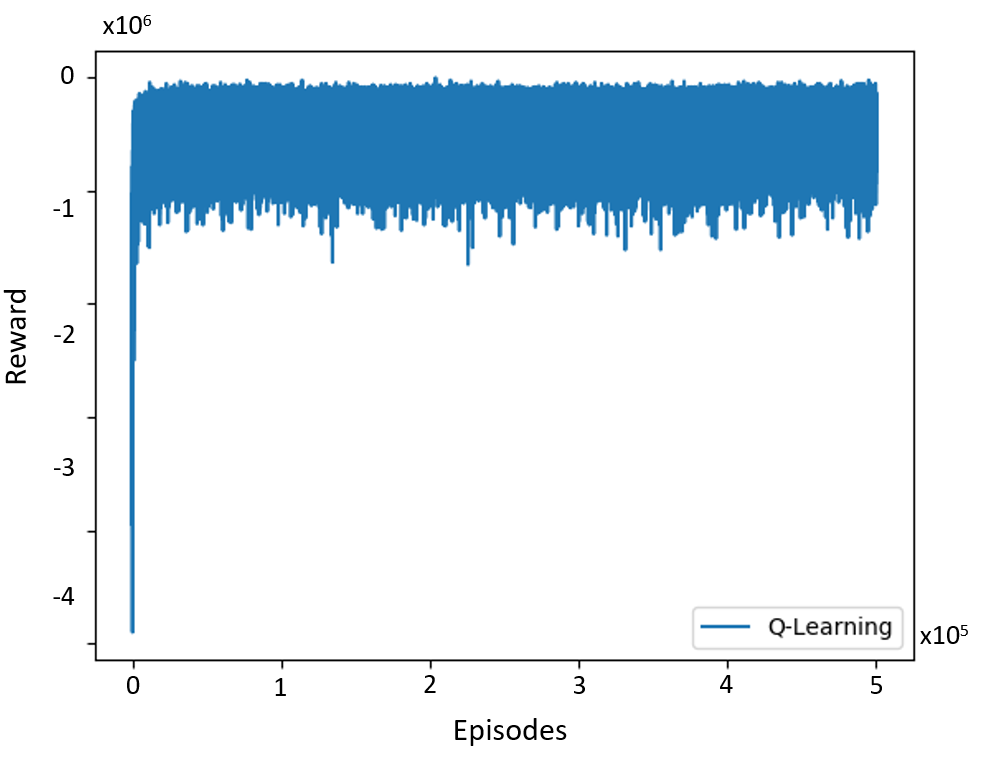}
            \caption{Learning curve for second alternative RL model. \label{fig:LearningCurveAltModel2}}
          \end{subfigure}
          \end{tabular}
    \end{tabular}
  \caption{RL model results for three different training session. \label{fig:RLModelResults}}
\end{figure}

The number of jobs had to be reduced in order for the RL to converge to a viable, though nonoptimal, solution. This nonoptimality is highlighted by the fact that the LSMS was not chosen to complete $M_2$ despite it being the better choice as described in Fig. \ref{fig:RLResults}. Fig. \ref{fig:ActionStateSpaceRL} shows that the model has converged and with additional training time the chosen actions for the given states will not change. It is important to note that the RL did not always converge to a correct schedule. Fig. \ref{fig:RLModelResults} shows the variance between state-action spaces. This is most likely due to the fact that the state-action space is very large compared to the number of correct state-actions it needs to learn for an optimal schedule.

\section{Discussion \& Conclusion} \label{sec:DiscussionConclusion}

The work presented here describes a novel application of FJSP to frame the in-space autonomous assembly problem providing a general description that can then be utilized by different solution formulations. The proposed MIP solution efficiently solved to optimality for the test instances. This approach solves the deterministic version of the problem from an offline (predetermined solution) planning perspective. This solution is ideal since it guarantees the most efficient way to complete the project. However, in this formulation, all of the precedence and holding constraints had to be directly encoded in the constraint equations. As autonomous assembly scenarios become more complex, the interjob dynamics will become harder to explicitly define. In contrast to the MIP results, the RL approach did not successfully converge to an optimal schedule. While it did learn the interjob dynamics to create a policy of decisions based on interactions with the environment thus allowing for flexible scheduling based on unforeseen circumstances, it was limited by the state space formulation. Future research will evaluate ways to combine the strengths of these two methods along with understanding stochastic elements of uncertainty to give autonomous systems the ability to autonomously learn and solve for schedules to in-space assembly projects facilitating persistent space exploration.

\bibliography{citations}
\bibliographystyle{ieeetr}



\end{document}